# Is There a Role for Qualitative Risk Assessment?

Paul Krause
Imperial Cancer Research Fund
London
WC2A 3PX

John Fox
Imperial Cancer Research Fund
London
WC2A 3PX

Philip Judson
Heather Lea
Norwood
Harrogate
HG3 1TE

## Abstract

Classically, risk is characterised by a point value probability indicating the likelihood of occurrence of an adverse effect. However, there are domains where the attainability of objective numerical risk characterisations is increasingly being questioned. This paper reviews the arguments in favour of extending classical techniques of risk assessment to incorporate meaningful qualitative and weak quantitative risk characterisations. A technique in which linguistic uncertainty terms are defined in terms of patterns of argument is then proposed. The technique is demonstrated using a prototype computer-based system for predicting the carcinogenic risk due to novel chemical compounds.

## 1 INTRODUCTION[1]

In the complex and dynamic world in which we live, risk assessment is taking an increasingly important role in both public and private decision and policy making. Decisions made on the basis of the possibility of global warming, for example, may have far reaching financial, environmental and sociological consequences. Equally, an inability to persuade a local authority to accept a subjective assessment of risk due to a "dangerous" road may have serious and tragic personal consequences. These two examples have been deliberately chosen as they can both be used to illustrate the extreme difficulty of providing a reliable point value measure of the likelihood of realisation of a perceived hazard. Nevertheless, the potential adverse consequences are so great that some meaningful risk characterisation is needed to enable a coherent decision to be made on the appropriate action.

This paper explores the problem of risk assessment in a domain, chemical carcinogenicity, for which quantitative risk assessment has been widely and publicly questioned (Carter, 1991). Some of the arguments against quantitative risk assessment in certain domains will be rehearsed in the next section. It should be emphasised that the argument is not against the use of quantitative risk assessment per se. Rather, that there are situations where traditional methods of risk assessment need extending to enable weak quantitative, or even purely qualitative statements of risk to be made. A specific interest of the authors is the development of a computer-based assistant for the assessment of potential carcinogenic risk of novel chemical compounds. An early prototype of this system will be used as a focus for discussing some approaches that have been taken to the qualitative and weak quantitative assessment of risk.

This paper does not claim to provide a full solution to the problem of qualitative risk assessment, although the results obtained to date are promising. The main reason for writing this paper at this stage is that this is a very important issue and needs to be raised as a matter of urgency. It is an unfortunate fact of life that information relating to many of the risks we face in our daily lives is often sparse and incomplete. If the uncertainty community could bend its collective mind to providing techniques for effective risk assessment and communication in such domains, this would be a major contribution to society as a whole.

## 2 THE CASE FOR NON-NUMERICAL RISK ASSESSMENT - NOT ENOUGH DEATHS

For certain technologies, such as electronic systems, civil-engineering structures and mechanical systems, established statistical models are available for making precise and reliable estimates of the likelihood of system failure. Significant quantities of historical data on the failure rates of standard components may be available, for example, or it may be possible to generate reliable simulations of system behaviour. Nevertheless, the contention that an objective, scientific assessment of risk is an achievable goal *is* being questioned. A 1992 report on *Risk Assessment* published by Britain's Royal Society included the comment that

> the view that a separation can be maintained between "objective" risk and "subjective" or perceived risk has come under increasing attack, to the extent that it is no longer a mainstream position. (Roy. Soc., 1992)

That is, there is no such thing as "objective" risk. Rather, risk is culturally constructed (see the next paragraph for an example of this). In fact, this quotation rather overstates the case in the context of the mainstream research and literature on safety and risk management. For example, Brit-ain's Department of Transport draws a firm distinction between "actual" danger (objective risk) and perceived danger (subjective risk). Their position is that if a road does not have a fatality rate above a certain threshold which is considered normal, and therefore acceptable, it will not be eligible for funds for measures to improve safety ("normal" is about 1.2 fatalities per 100 million vehicle kilometres). Nevertheless, their position can lead to conflict. Consider the following scenario.

---

[1] An extended version of this paper, with larger screen images, is available via anonymous ftp from acl.icnet.uk in ~ftp/pub/StAR/qual_risk.ps.

Increasing traffic over the years leads to a straight road through a residential area being considered "dangerous" by local residents. They plead with the local authority for something to be done to calm the traffic. In the meantime, children are warned to stay away from the road if possible, and to take extreme care if they do need to cross the road for any reason. As a result, the fatality rate on that road stays low - people are taking extra care not to expose themselves to a situation which they *perceive* as potentially risky. The local authority has no observable *measure* of increased risk, so nothing is done. Then a tragic accident does take place. Amidst public outcry, the local authority promises that traffic calming measures will be in place at the earliest opportunity. Which was the *real* risk: the perceived risk of the residents, or the objective risk (estimate) of the local authority?

Sadly, this is not an academic exercise. In 1991, Britain's Permanent Secretary for the Department of Transport announced that "funds for traffic calming will be judged on casualty savings, not environmental improvements or anxiety relief" (Brown, 1991). This extreme position has lead to many conflicts between local people and the Department of Transport, very tragically with consequences not dissimilar to the above hypothetical scenario. More recently the situation has changed slightly, but the whole issue of the distinction between one person's perceived risk and another's objective (estimate of) risk remains a major source of conflict in many fields (Adams, 1995). Although it is not possible to quantify the subjective perceptions of risk, statistics are available on the risk management activities that are taken in response to these subjective judgements:

> In 1971, for example, 80 per cent of seven and eight year old children in England travelled to school on their own, unaccompanied by an adult. By 1990 this figure had dropped to 9 per cent; the questionnaire survey disclosed that the parents' main reason for not allowing their children to travel independently was fear of traffic. (Adams, 1995, p.13)

The above provides an example in which an authority has an objective estimate of risk which can be expressed as a single number (fatalities per so-many vehicle kilometres). This is questioned by a public which has a quite different perception of risk which they find much harder to express, other than through an extended debate with the authority. There are situations, however, where the authorities themselves cannot agree on an objective measure of risk. One such is the assessment of carcinogenicity due to chemicals. This will be the focus of attention for the remainder of this paper.

> Of over five million known chemical substances, only thirty are definitely linked with cancer in humans, and only 7,000 have been tested for carcinogenicity: the rest is darkness. (Adams, 1995)

There are a number of approaches which one might take to assessing the carcinogenic risk due to a certain chemical. The surest way of obtaining a reliable, objective risk estimate is to establish a causal mechanism and/or obtain direct statistical data on a population of humans that have been exposed to the chemical. Clearly, it would be unacceptable to subject a population to a controlled release of the chemical, but epidemiological data are sometimes available on a population which is known to have been exposed in the past to a chemical (cigarette smoking is a case in point). However, as the quote from John Adams indicates, such information is available for very few chemicals. By far the most frequently used techniques for trying to establish the carcinogenic risk due to chemicals involve *in vivo* or *in vitro* tests. In the case of *in vivo* tests, regular doses of the substance under study are delivered to a population of test animals. The relative increase in tumour incidence with respect to a control population, as a function of dosage, is then used as the basis for an estimate of the increased risk to humans through exposure to the same substance. This estimate requires, however, a number of extrapolations to be made. In order for the experiment to be carried out within an acceptably short period of time, the test animals are subjected to a much higher dose rate than would be anticipated in the human population, so the first extrapolation is from a high dose rate to a low dose rate. Secondly, the results must be extrapolated across species from a population of test animals to humans.

The nature and validity of both forms of extrapolation are the subject of wide ranging disagreements between experts. Figure 1, for example, shows a number of alternative dose-response extrapolations from the same empirical data. It can be seen that the resulting risk predictions at low dose rates vary by many orders of magnitude. The extrapolation between animals and humans is subject to an equally high degree of disagreement. In an extreme case, suppose a chemical is seen to induce an increased incidence of tumours on a gland in rats that has no analogue in humans. What conclusions can be realistically made about that chemical's potential carcinogenicity in humans? Even without considering the moral position associated with such experiments, serious questions can be asked about the ultimate value of such tests as predictors of human carcinogenic risk.

In the case of *in vitro* tests, such as the Ames Test (Ashby & Tennant, 1991), a "test tube" experiment is carried out to see if the chemical under study can induce mutations in DNA. However, although genetic mutation is a *necessary* prerequisite to the development of a cancer, mutagenicity is not a *sufficient* criterion for carcinogenisis. In addition, a "carcinogen" may not directly cause the genetic mutation that induces a cancer; it may play a role in facilitating rather than directly causing the genetic damage. Therefore, the results of *in vitro* tests as predictors of carcinogenic risk are also open to a wide range of interpretations (e.g.

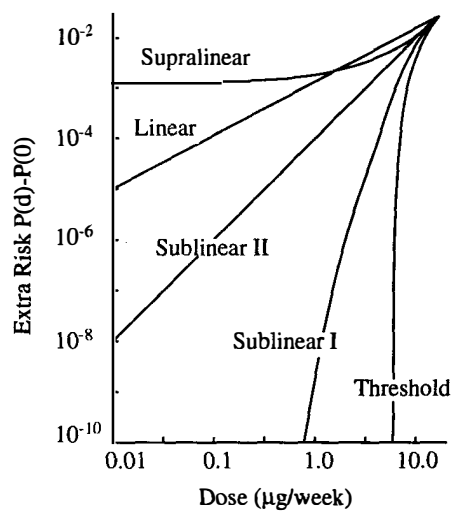

Figure 1: Results of alternative dose-response extrapolations from the same empirical data (after National Research Council, 1992). *Extra Risk* is the increase in probability of occurrence of an adverse effect, as a function of dose, above the probability of occurrence of that effect in the population as a whole.

Ashby, 1994).

In this domain questioning the value of "objective" point value measures of risk really *is* the mainstream position. The report of the UK Department of Health's Committee on Carcinogenicity of Chemicals in Food (Carter, 1992) concludes:

> The committee does not support the routine use of quantitative risk assessment for chemical carcinogens. This is because the present models are not validated, are often based on incomplete or inappropriate data, are derived more from mathematical assumptions than from a knowledge of biological mechanisms and, at least at present, demonstrate a disturbingly wide variation in risk estimates depending on the model adopted.

Nevertheless, there are still situations when a meaningful risk characterisation is needed. The next section will introduce the StAR risk assessment project, and then the paper will continue with a discussion of the approaches to extending the scope of risk assessment which are being developed within that project.

## 3   GOALS OF *StAR*[1]

The approach to risk assessment which will be reported in this paper is being developed in the project *StAR* (for *St*andardised *A*rgument *R*eport). The top level goals of this project are:

* to develop a computer-based aid to risk assessment which can systematically accommodate diverse types of evidence, both qualitative and quantitative;

---
[1] The StAR consortium consists of: Judson Consulting; Imperial Cancer Research Fund, London; LHASA UK, School of Chemistry, Leeds University; Logic Programming Associates (LPA), London; Psychology Department, City University, London.

* to develop techniques for risk communication which are both comprehensive and intelligible.

*StAR* draws on earlier work on the development of a model for reasoning under uncertainty based on a simple intuitive notion of constructing arguments "for" and "against" a hypothesis (Fox et al., 1992). The net support for the hypothesis may then be evaluated using one of a number of techniques; the choice being dependent on the nature of the evidence available. Full details of the syntax and semantics of the formal model of argumentation may be found in (Krause et al., 1995a).

This project is to a great extent problem driven. A concrete goal is to develop a risk adviser to support the prediction of carcinogenic risk due to novel chemical compounds. The problems of toxicological risk assessment in general have been introduced above, and are discussed in more detail in (Krause et al., 1995b). The following points summarise both these discussions:

* toxicological risk assessments for chemicals may at best cover a very wide range of possible values;

* point value estimates conceal the uncertainties inherent in risk estimates;

* judgements based on the comparison of point values may be quite different from those based on the comparison of ranges of possible values;

* in very many cases the spread of possible values for a given risk assessment may be so great that a numerical risk assessment is completely meaningless.

Our aim is to allow the incorporation of numerical data where available, and to allow a grading of risk characterisation from quantitative through semi- or weak-quantitative to qualitative, contingent on the reliability and accuracy of the data available. The purely qualitative risk characterisation is the most controversial of these, and is in most need of discussion. Hence, this will form the focus of the major part of the remainder of this paper.

## 4   QUALITATIVE TERMS FOR RISK ASSESSMENT

The need for some form of qualitative risk characterisation has long been accepted by the U.S. International Agency for Research on Cancer (IARC) and Environmental Protection Agency (EPA). Their joint proposal for a qualitative classification scheme will be discussed in this section. This will be contrasted with a more general proposal for "symbolic uncertainty".

Both of the following approaches use terms whose semantics is defined in terms of logical states. This contrasts fundamentally with most of the existing work on the use of linguistic uncertainty terms, in which the underlying semantics is assumed to be probabilistic (e.g. Budescu & Wallsten, 1995).

IARC Classification System.

This is based on the U.S. EPA classification scheme. It uses a small set of terms which are defined to represent the current state of evidence. They take as their basis a classification of the weight of evidence for carcinogenicity into five groups for each of human studies and animal studies. These classifications are summarised here, but are defined precisely in (U.S. EPA, 1986).

For human studies, the classifications are:
*Sufficient evidence* of carcinogenicity. There is an established causal relationship between the agent and human cancer.
*Limited evidence* of carcinogenicity. A causal relationship is credible but not established.
*Inadequate evidence*. Few available data, or data unable to support the hypothesis of a causal relationship.
*No data.* Self-explanatory.
*No evidence.* No association was found between exposure and increased incidence of cancer in well-conducted epidemiological studies.

Similar classifications can be defined for animal studies. Data obtained from these classifications are then used to provide an overall categorization of the weight of evidence for human carcinogenicity (again, full definitions can be found in U.S. EPA, 1986):

*Known* Human Carcinogen:
*Sufficient* evidence from human (epidemiological) studies.

*Probable* Human Carcinogen:
*Sufficient* animal evidence and evidence of human carcinogenicity, or *at least limited* evidence from human (epidemiological) studies.

*Possible* Human Carcinogen:
*Sufficient* animal evidence but *inadequate* human evidence, or *limited* evidence from human studies in the *absence of sufficient* animal evidence.

*Not Classifiable*:
*Inadequate* animal evidence and *inadequate* human evidence, but *sufficient* evidence of carcinogenicity in experimental animals.

*Non carcinogenic* to Humans:
Evidence for lack of carcinogenicity.

Note that these uncertainty terms are defined specifically in the context of carcinogenicity risk assessment. Our criticism of them is primarily on this basis. It would be more useful to see a set of terms which were defined at a higher level of abstraction. This would enable their intention to be communicable to a person who was not necessarily familiar with the details of risk assessment in a specific domain. It would also enable their usage to be standardised across a wide range of domains.

Elvang-Gøransson et al's "logical uncertainty" terms.

An alternative set of terms with a precise mathematical characterisation was defined in (Elvang-Gøransson et al., 1993). These terms take the notion of logical provability as primitive. They then express successively increasing degrees of "acceptability" of the arguments which support the propositions of interest; as one progresses down the list there is a decrease in the tension between arguments for and against, a hypothesis P. A precise characterisation of these terms is quite lengthy, and so is not reproduced here. Full details and a discussion of their properties can be found in (Krause et al., 1995a). The following is intended to give a reasonably intuitive informal description.

P is *open*
if it is *any* well-formed formula in the language of the logic (one may be unable to construct any arguments concerning it, however).

P is *supported*
if an argument, possibly using inconsistent data, can be constructed for it.

P is *plausible*
if a consistent argument can be constructed for it (one may also be able to construct a consistent argument against it).

P is *probable*
if a consistent argument can be constructed for it, and no consistent argument can be constructed against it.

P is *confirmed*
if it satisfies the conditions of being probable and, in addition, no consistent arguments can be constructed against any of the premises used in its supporting argument.

P is *certain*
if it is a tautology of the logic. This means that its validity is not contingent on any data in the knowledge-base.

No quantitative information is used in the definition of these terms; they use purely logical constructions. However, it should be clear that they allow a unidimensional scaling.

A problem still remains. Although these terms do have a precise definition, it is an open question whether they have "cognitive validity" (see the next section). If not, then they will be open to misinterpretation as a vehicle for communication.

## 5   THE *StAR* DEMONSTRATOR

An alternative strategy to either of the above is to see if it is possible to establish specific patterns of argument as qualitative landmarks. The aim is then to associate those patterns with linguistic terms in a way which has "cognitive validity"; that is, where the definitions reflect in some way people's intuitive usage of the associated terms. In order to explore these ideas further, a risk assessment demonstrator has been built which uses a small set of linguistic terms as part of the reporting facility. A brief run-through of the demonstrator will be used in this section to illustrate the general approach. Some more detailed definitions of argument structures will be given in the next section. Note that the linguistic terms used in this section are for illustration only, the precise choice of terms being the subject of ongoing work.

The demonstrator is a prototype for a computer based assistant for the prediction of the potential carcinogenic risk due to novel chemical compounds. A notion of hazard identification is taken as a preliminary stage in the assessment of risk. The hazard identification used here draws heavily on the approach taken in the expert system DEREK, which is used for the qualitative prediction of possible toxic action of chemical compounds (Sanderson & Earnshaw, 1991). DEREK is able to detect chemical sub-structures within molecules, known as structural alerts, and relate these to a rule-base linking them with likely types of toxicity. In the demonstration, the structural alerts have been taken from a U.S. FDA report identifying sub-structures associated with various forms of carcinogenic activity (U.S. FDA, 1986).

The user of the carcinogenicity risk adviser presents the system with the chemical structure of the compound to be assessed, together with any additional information which may be thought relevant (such as possible exposure routes, or species of animal that will be exposed to the chemical). The chemical structure may be presented using a graphical interface.

The database of structural alerts is then searched for matches against the entered structure. If a match is found, a theorem prover tries to construct arguments for or against the hazard being manifest in the context under consideration. Having constructed all the relevant arguments, a report is generated on the basis of the available evidence.

For ease of presentation, the examples use a very simplified database, and some of the following assessments may be chemically or biologically naive.

In the first example, the structure for aniline has been drawn. This is normally regarded as a carcinogen. However, in this instance, the interest is in assessing the chemical in the context of guinea pigs being exposed to it.

In this case, the following report is generated.

The system has concluded that the case for the substance being carcinogenic is "equivocal". A hazard alert for a proximate carcinogen has been found (the structure is an aromatic amine), but there is evidence both for (the chemical has a logP value which indicates that it will be easily absorbed into fatty tissue), and against (the chemical is unlikely to be metabolised into a direct acting carcinogen) realisation of the hazard. To simplify the construction of the report, the arguments are summarised by simply listing the grounds, basic facts, used in constructing each argument. Work is underway to provide a more easily assimilable summarisation of the argument at this stage. The user may click on either argument to obtain a more detailed explanation.

Had the value provided for logP been below 2.0, then the following report would have been generated.

It will be noticed that these two reports also contain some numerical information as part of the risk report. The argumentation model permits a measure of confidence in a hypothesis which is equivalent to the Dempster-Shafer probability of provability (Krause et al., 1995a). If numerical coefficients are available for all the relevant axioms, then measures of belief and plausibility can be given for the risk statement. It can be seen that the qualitative behaviour of these coefficients follows the linguistic term used in the report (the "equivocal" report has Bel=0.68, Pl=0.76; the "improbable" report has Bel=0.00, Pl=0.24). The linguistic terms are not, however, defined in terms of fixed probability intervals.

Issues arise about the precise interpretation of these numbers, however, and these will be discussed in section 7.

## 6  ARGUMENTS AND CASES

The previous section introduced the concept of matching linguistic uncertainty terms to structures of arguments. The thesis is that certain useful structures can be identified as "landmarks" in the space of all possible argument structures, and that natural language is capable of capturing the distinctions between these landmarks (Fox, 1986; Fox & Krause, 1991).

Some example definitions of such landmarks will be given in this section. A more extensive set of definitions can be found in (Fox & Krause, 1994). Experiments are currently under way to see if linguistic terms can be consistently associated in a meaningful way with such patterns of argument.

As a starting point four distinct "epistemic" states for a proposition $p$ can usefully be distinguished: p is con-firmed; p is supported; p is opposed; p is excluded. It is useful to agree a notation to distinguish succinctly between these states. The following has been used in earlier work (Fox et al., 1992):

p:+ ≡ "p is supported"
p:− ≡ "p is opposed"
p:++ ≡ "p is confirmed"
p:—— ≡ "p is excluded"

## 6.1 ARGUMENTS

We wish to raise arguments to the level of "first class objects" so that we may reason about *them*, as well as about the propositions of interest. A way of achieving this is to actually identify those facts and rules which are used to construct the argument:

**Definition 6.1.1**

If $\delta \subseteq \Delta$ is minimal such that $\delta \vdash p$, then $(\delta,p)$ is an argument for p from $\Delta$. $\delta$ itself is referred to as the *grounds* of the argument.

If we now add in the above notation, we are able to distinguish four different basic classes of argument:

If $\delta \vdash p{:}+$, then $(\delta,p)$ is a supporting argument;
If $\delta \vdash p{:}-$, then $(\delta,p)$ is an opposing argument;
If $\delta \vdash p{:}{++}$, then $(\delta,p)$ is a confirming argument;
If $\delta \vdash p{:}{-\!-}$, then $(\delta,p)$ is an excluding argument.

## 6.2 CASES

The set of all arguments which impact on a single proposition, constitutes the case concerning that proposition:

**Definition 6.2.1**

For a given proposition p

$\{(\delta,p) : \exists q \in \{++, +, -, --\} \bullet \delta \vdash p{:}q\}$ is the *"case concerning p"*.

Two sub-cases of an overall case can be usefully distinguished:

**Definition 6.2.2**

For a given proposition p

$\{(\delta,p) : \exists q \in \{++, +\} \bullet \delta \vdash p{:}q\}$ is the *"case for p"*;

$\{(\delta,p) : \exists q \in \{-, --\} \bullet \delta \vdash p{:}q\}$ is the *"case against p"*.

By altering the conditions on the kind of arguments, we may be able to define other useful sub-cases of an overall case.

## 6.3 CLASSES OF CASES

One further item of notation needs to be introduced now.

**Definition 6.3.1**

$|(\delta,p)|$ is the "weight" of the argument $(\delta,p)$.

$|\{(\delta,p) : conditions\}|$ is the aggregate weight of those arguments $(\delta,p)$ satisfying *conditions*.

The latter may be just a "head count". This will usually be indicated by comparing the weight to some integer. For example $|\{(\delta,p) : \exists q \in \{++, +\} \bullet \delta \vdash p{:}q\}| \geq 1$ might be paraphrased as "there are one or more arguments for p". In contrast

$|\{(\delta,p) : \exists q \in \{++, +\} \bullet \delta \vdash p{:}q\}|$
$\qquad > \quad |\{(\delta,p) : \exists q \in \{-, --\} \bullet \delta \vdash p{:}q\}|$

means the aggregate weight of the *case for* is greater than the *case against* (be it a head count, probabilistic aggregation, or whatever).

Patterns identifying "classes of cases" can now be defined. Here are some examples. The first is a general pattern for all those cases where there are both arguments for and arguments against, the second where there are arguments for, but at least one excluding argument, and so on. For each pattern, an English language gloss precedes the formal definition.

For a given proposition p:
There is both a case for and a case against p. The term "equivocal" was used to describe this structure in the demonstrator.

$|\{(\delta,p) : \exists q \in \{++, +\} \bullet \delta \vdash p{:}q\}| > 0$
$\qquad \& \quad |\{(\delta,p) : \exists q \in \{-, --\} \bullet \delta \vdash p{:}q\}| > 0$

There is a case for p, but at least one excluding argument
$|\{(\delta,p) : \exists q \in \{++, +\} \bullet \delta \vdash p{:}q\}| > 0$
$\qquad \& \quad |\{(\delta,p) : \delta \vdash p{:}{-\!-}\}| > 0$

p has been confirmed, but there is still a case against p
$|\{(\delta,p) : \delta \vdash p{:}{++}\}| > 0$
$\qquad \& \quad |\{(\delta,p) : \exists q \in \{-, --\} \bullet \delta \vdash p{:}q\}| > 0$

p has been both confirmed and excluded
$|\{(\delta,p) : \delta \vdash p{:}{++}\}| > 0$
$\qquad \& \quad |\{(\delta,p) : \delta \vdash p{:}{-\!-}\}| > 0$

The last pattern means that it is possible to derive a contradiction in the classical sense. We can write the argument for the contradiction as $(\delta, \bot)$.

In a similar way, groups of patterns can be defined covering the situations where: either there is no case against, or there are at least no excluding arguments; there is either no case for, or at least there are no confirming arguments; and where purely negative statements are made about the state of evidence (Fox & Krause, 1994). It is not clear whether all of these will be useful, however.

Apart from distinguishing between support and confirmation (or opposition and exclusion), the above patterns do not make any distinctions with respect to the weight of evidence. The following basic distinctions can be made.

For single arguments $(\delta,p)$ and $(\gamma,p)$, where $\delta \vdash p{:}+$ and $\gamma \vdash p{:}-$

A supporting and an opposing argument are of equal weight
$$|(\delta,p)| = |(\gamma,p)|$$

The supporting argument is stronger than the opposing argument
$$|(\delta,p)| > |(\gamma,p)|$$

The supporting argument is weaker than the opposing argument

$$|(\delta,p)| < |(\gamma,p)|$$

Analogous patterns can be defined for sets of arguments.

Finally, there is the possibility that arguments themselves may be attacked. There is scope here for quite complex interactions between arguments; attacking arguments, attacking arguments that attack arguments, and so on.

The above just gives a selection of argument structures that *may* be useful as qualitative landmarks. To continue to develop the thesis outlined at the beginning of this section, we need to answer two questions. Which of these landmarks are recognised as important? Of those that are important in evaluating states of evidence, what is the language people use to recognise them?

Experiments are being designed to see if it is possible for subjects to classify examples of these abstract definitions consistently into categories where the members of each category consist of arguments conforming to the same, or a similar, abstract definition. If this is achievable, then the experimental work will move on to study whether linguistic terms can be meaningfully and consistently associated with these categories.

# 7   DISCUSSION

There is an important difference between the system of argumentation that has just been presented, and the non-monotonic models that incorporate mechanisms of defeat and rebuttal such as Poole (1985), Nute (1988), Loui (1987) and the various default logics. In the non-monotonic argumentation models, defeat and rebuttal may result in conclusions being retracted from the current belief state. This is not the case here. The argument structure is presented to the user, and the evidence state summarised by qualifying the conclusion with a linguistic term. This term indicates the general class of argument structures to which the specific case belongs. It is then up to the users of the system to be more, or less, cautious about which conclusions they accept, rather than the designer of the argument system.

This has important implications for risk communication. Even if the risk of some adverse effect has been strongly discounted, the risk is still presented together with the justification(s) for discounting that risk. The intent is that the risk characterisation should be transparent to the recipient, and he or she acts as final arbiter. The major benefit of this is that there is an explicit representation of the state of evidence concerning a proposition. However, the provision of a decision rule to aid the user in acting on the basis of the qualitative risk characterisation still needs to be addressed.

This paper focuses on qualitative risk assessment because it is the aspect of the StAR project that is most in need of discussion. However, the aim is to incorporate weak-quantitative and quantitative risk characterisations where possible. As indicated in section 5, some questions arise from this.

Where numerical coefficients have been associated with the axioms in the demonstrator's database, their precise value was to a certain extent arbitrary as their use here is merely intended to be illustrative. Nevertheless, an important question is raised. In this context, what precisely should the numbers mean? Remember that the system is intended to give some form of risk *prediction*, drawing on some prior general knowledge about indicators for potential carcinogenic activity. Should a numerical value, or interval of possible values indicate:

- the lifetime likelihood of someone being exposed to the chemical developing cancer;

or

- a subjective estimate of belief in the statement "this chemical is carcinogenic"?

Given the difficulty of assessing risk due to chemicals that have been subject to some experimental study, discussed in section 2, it does not seem realistic to suppose that meaningful probabilities could be assigned to the general indicators of risk that are used in the *StAR* database. However, it *may* be possible to elicit subjective indications of the relative plausibility of the conclusions drawn from the generic knowledge that is elicited for the *StAR* database. So, whilst 1) above is unlikely to be achievable, an ability to include a value or range of values that conform to 2) *may* be realistic, at least in certain circumstances.

The *StAR* database can be viewed as consisting of facts (about the structure of the chemical of interest, its physical properties and the context in which it will be used) together with generic knowledge about possible mechanisms of carcinogenicity (chemical sub-structures with believed associations with carcinogenic activity, possible metabolic pathways, exposure routes, and so on). From this knowledge, *StAR* draws *plausible* conclusions. It may well be that it would be more appropriate to model the generic knowledge as conditional assertions (Dubois and Prade, 1994) rather than by the more naive approach of rules with certainty coefficients as used here. In fact, the definition of an argument used in this paper is consistent with that of Benferhat, Dubois and Prade (Benferhat et al., 1993). Hence a possibilistic encoding of rational inference may be an appropriate unifying framework for the qualitative and quantitative aspects of the risk characterisation, in the context where the numerical coefficient indicates a measure of plausibility of the conclusion.

A crucial issue is still outstanding, however. Although this paper offers some ideas for qualitative and (semi-)quantitative risk characterisations in the absence of reliable statistical data, there still remains the question of how to act on the basis of such a risk assessment. The implicit claim is that there is scope for fairer decisions to be made if a risk assessment is carried out in an argumentation framework, because:

- the reasoning behind the risk assessment is readily open to inspection;
- a party is able to counter a numerical risk assessment with a subjective, but clearly structured, risk characterisation.

In the case of the carcinogenic risk application being developed in StAR, we do not see the decisions that are made after consultation with the system as being along the lines of "we shall, or shall not allow this chemical into the public arena". Rather, it is intended to assist in the prioritisation of further investigations into the chemical of interest. That is, the decisions in question are of a different type from the rational choice of alternatives that is classically the domain of decision theory.

Apart from the specific application, it is hoped that the work described on providing a framework for the structuring of cases will contribute to a more orderly debate between two parties with differing conceptions of the risk associated with a specific situation.

## 8 CONCLUSION

This paper reports work on the development of qualitative techniques for risk assessment. The main aim of the paper is to bring the motivation for the work into the public arena. The main points are:
- there are situations in which a numerical assessment of uncertainty is unrealistic,

yet;
- some structured presentation of the case that supports a particular assessment may still be required, and is achievable.

The concept of characterising risk by linguistic terms defined by relating them to patterns of argument has been demonstrated. This was then followed up with more detailed definitions of categories of cases (collections of arguments) which are being used as the basis for further studies.

This is not just an academic exercise in the use of argument structures. A solution to the problems outlined in the first part of this paper is of vital importance. It is for this reason that we wish to bring discussion of this work into the public arena at this stage.

### Acknowledgements

The valued assistance of Saki Hajnal in the final preparation of this paper is gratefully acknowledged. This work is being supported under the DTI/EPSRC Intelligent Systems Integration Programme, project IED/4/1/8029.